\documentclass{article} 
\usepackage{iclr2024_conference,times}

\usepackage{hyperref}
\usepackage{url}

\usepackage[group-separator={,}]{siunitx}
\usepackage{booktabs} 
\usepackage{color}
\usepackage{listings}
\usepackage{enumitem}
\usepackage{threeparttable}
\usepackage{multirow}
\usepackage{hhline}
\usepackage{subcaption}
\usepackage{array}
\usepackage{pifont}
\usepackage{booktabs}
\usepackage{dsfont}
\usepackage{microtype}
\usepackage{stmaryrd}
\usepackage{amsmath}

\usepackage{amsthm} 
\usepackage{url}
\usepackage{bbm}
\usepackage{xspace}
\usepackage[normalem]{ulem}
\usepackage{booktabs, siunitx}
\usepackage{xcolor}
\usepackage{tkz-euclide}
\usepackage[framemethod=tikz]{mdframed}
\usepackage{empheq}
\usepackage[most]{tcolorbox}
\usepackage{adjustbox}
\usepackage{wrapfig}
\usepackage{algorithm}
\usepackage{algorithmic}
\usepackage{pgfplots}
\usepgfplotslibrary{groupplots}
\usepackage{pgfplotstable}

\usepackage{balance}
\expandafter\def\expandafter\UrlBreaks\expandafter{\UrlBreaks
  \do\a\do\b\do\c\do\d\do\e\do\f\do\g\do\h\do\i\do\j%
  \do\k\do\l\do\m\do\n\do\o\do\p\do\q\do\r\do\s\do\t%
  \do\u\do\v\do\w\do\x\do\y\do\z\do\A\do\B\do\C\do\D%
  \do\E\do\F\do\G\do\H\do\I\do\J\do\K\do\L\do\M\do\N%
  \do\O\do\P\do\Q\do\R\do\S\do\T\do\U\do\V\do\W\do\X%
  \do\Y\do\Z}

\usepackage{bbold} 

\expandafter\def\expandafter\UrlBreaks\expandafter{\UrlBreaks
  \do\a\do\b\do\c\do\d\do\e\do\f\do\g\do\h\do\i\do\j%
  \do\k\do\l\do\m\do\n\do\o\do\p\do\q\do\r\do\s\do\t%
  \do\u\do\v\do\w\do\x\do\y\do\z\do\A\do\B\do\C\do\D%
  \do\E\do\F\do\G\do\H\do\I\do\J\do\K\do\L\do\M\do\N%
  \do\O\do\P\do\Q\do\R\do\S\do\T\do\U\do\V\do\W\do\X%
  \do\Y\do\Z}

\definecolor{codegreen}{rgb}{0,0.6,0}
\definecolor{codegray}{rgb}{0.5,0.5,0.5}
\definecolor{codepurple}{rgb}{0.58,0,0.82}
\definecolor{backcolour}{rgb}{1,1,1}
 
\lstdefinestyle{mystyle}{
    backgroundcolor=\color{backcolour},   
    commentstyle=\color{codegreen},
    keywordstyle=\color{magenta},
    numberstyle=\tiny\color{codegray},
    stringstyle=\color{codepurple},
    basicstyle=\ttfamily\footnotesize,
    breakatwhitespace=false,         
    breaklines=true,                 
    captionpos=b,                    
    keepspaces=true,                 
    numbers=left,                    
    numbersep=5pt,                  
    showspaces=false,                
    showstringspaces=false,
    showtabs=false,           
    tabsize=2,
    xleftmargin=1.5em
}

\newtheorem{theorem}{Theorem}[section]

\newtheorem{definition}{Definition}[section]
\newtheorem{example}{Example}[section]

\newcommand{\technique}{VeriTraCER\xspace}
\newcommand{\ourabsname}{Simul-CROWN\xspace}
\newcommand{\ourabs}{Ours}
\newcommand{\deltarobust}{$\delta$-robustness\xspace}
\newcommand{\nfoldrobust}{cross-model validity\xspace}

\newcommand{\student}{OULA\xspace} 

\newcommand{\bfx}{\mathbf{x}}
\newcommand{\cfx}[1]{#1'}
\newcommand{\bfz}{\theta} 
\newcommand{\generalzb}{\mathbf{z}}

\newcommand{\x}{\bfx}

\newcommand{\param}[1]{\theta_{#1}}

\newcommand{\gen}{g}

\newcommand{\rs}[0]{f_\mathrm{m}}
\newcommand{\spred}[1]{\bar{#1}}
\newcommand{\robustratio}{\kappa}

\newcommand{\lbalpha}{\boldsymbol{\alpha}^l}
\newcommand{\ubalpha}{\boldsymbol{\alpha}^u}
\newcommand{\lbbeta}{\beta^l}
\newcommand{\ubbeta}{\beta^u}
\newcommand{\lbmu}{\boldsymbol{\mu}^l}
\newcommand{\ubmu}{\boldsymbol{\mu}^u}
\newcommand{\lbnu}{\nu^l}
\newcommand{\ubnu}{\nu^u}

\newcommand{\mult}[0]{\mathcal{M}}
\newcommand{\loss}{\mathcal{L}}
\newcommand{\lossmse}{\loss_\mathrm{MSE}}
\newcommand{\lossacc}{\loss_{\mathrm{A}}}
\newcommand{\lossval}{\loss_{\mathrm{V}}}
\newcommand{\lossf}{\loss_f}
\newcommand{\lossgen}{\loss_{\gen}}
\newcommand{\lossqual}{\loss_{\mathrm{Q}}}
\newcommand{\lossrobust}{\loss_\mathrm{R}}

\newcommand{\abs}[1]{#1^{\sharp}}
\newcommand{\absibp}[1]{#1^{\sharp\mathrm{IBP}}}
\newcommand{\abscibp}[1]{#1^{\sharp\mathrm{CIBP}}}
\newcommand{\absours}[1]{#1^{\sharp\mathrm{\ourabs}}}
\newcommand{\lb}[1]{#1^{L}}
\newcommand{\ub}[1]{#1^{U}}
\newcommand{\lbs}[1]{#1^{l}}
\newcommand{\ubs}[1]{#1^{u}}
\newcommand{\interval}[1]{#1^\sharp}
\newcommand{\slope}{\boldsymbol{\alpha}}
\newcommand{\itcpt}{\beta}
\newcommand{\cfxslope}{\boldsymbol{\mu}}
\newcommand{\cfxitcpt}{\nu}

\renewcommand\algorithmiccomment[1]{%
  \hfill//\ {#1}%
}

\newcommand{\cf}[1]{CE#1\xspace}
\newcommand{\robustcfx}[0]{RobustCE\xspace}

\makeatletter
\renewcommand{\paragraph}{%
  \@startsection{paragraph}{4}%
  {\z@}{0.5ex \@plus 0.2ex \@minus 0.1ex}{-1em}%
  {\normalfont\normalsize\bfseries}%
}
\makeatother

\usepackage{amsmath}
\usepackage{amssymb} 
\usepackage{mathtools}

\usepackage[capitalize,noabbrev]{cleveref}

\theoremstyle{plain}
\theoremstyle{definition}

\theoremstyle{remark}

\usepackage[textsize=tiny]{todonotes}

\title{Verified Training for Counterfactual Explanation Robustness under Data Shift}

\author{Anna P. Meyer\thanks{Equal contribution}\ \ , Yuhao Zhang$^{*}$, Aws Albarghouthi \& Loris D'Antoni 
\\
Department of Computer Sciences\\
University of Wisconsin - Madison \\
Madison, WI, USA \\
\texttt{\{annameyer, yuhaoz, aws, loris\}@cs.wisc.edu} \\
}

\iclrfinalcopy 
\begin{document}

\maketitle

\begin{abstract}
Counterfactual explanations (\cf{s}) enhance the interpretability of machine learning models by describing what changes to an input are necessary to change its prediction to a desired class.  
These explanations are commonly used to guide users' actions, e.g., by describing how a user whose loan application was denied can be approved for a loan in the future. 
Existing approaches generate \cf{s} by focusing on a single, fixed model, and do not provide any formal guarantees on the \cf{s'} future validity.
When models are updated periodically to account for data shift, if the generated \cf{s} are not \emph{robust} to the shifts, users' actions may no longer have the desired impacts on their predictions.
%
This paper introduces \technique, an approach that \emph{jointly} trains a classifier and an explainer to explicitly consider the robustness of the generated \cf{s} to small model shifts. 
\technique optimizes over a carefully designed loss function that ensures the \emph{verifiable} robustness of \cf{s} to local model updates, thus providing deterministic guarantees to \cf{} validity.
Our empirical evaluation demonstrates that \technique{} generates \cf{s} that (1) are verifiably robust to small model updates and (2) display competitive robustness to state-of-the-art approaches in handling empirical model updates including random initialization, leave-one-out, and distribution shifts. 
\end{abstract}

\section{Introduction}
\label{sec:intro}

Machine learning
models are increasingly used to support decision-making in sectors such as banking, education, social services, and criminal justice. Due to the high stakes of these decision-making settings, and the fact that model internals are often both proprietary and too complex for humans to understand directly, laws such as the GDPR~\citep{gdpr} and ECOA~\citep{ecoa} require that explanations be offered to users who are subject to these models' predictions. 
Explanations commonly take the form of counterfactual explanations (\cf{s}), which serve as guidelines for how an input can change in order to receive a different decision in the future. For instance, an individual who is denied a loan may receive a \cf{}that says their application would have been accepted had their salary been \$\num{5000} higher. If the applicant wishes to obtain the loan, they can work to increase their salary and then reapply. 

However, machine learning models must be periodically updated to account for new data and avoid declining performance due to distribution shift. Even if the \cf{} is valid (i.e., produces the desired prediction) at the time it is generated, there is no guarantee that it will remain valid after one or more routine model updates. So, the individual who successfully raises their salary by the requested \$\num{5000} may still be rejected if they reapply for a loan 6 months later, because the model internals shifted. 

To preserve the validity of \cf{s} across model shifts, we want to generate \cf{s} that are \emph{robust} to small model changes. Existing work aims to generate robust \cf{s} by increasing the distance from the original input to its \cf{}~\citep{hamman2023robust}, finding \cf{s} in areas of low Lipschitz constants~\citep{black2021consistent}, or using a minmax objective in the \cf{}generation process~\citep{roar}. While all of these methods yield improved robustness over standard training, they fail to provide \emph{formal guarantees} on the \cf{s'} robustness, they require solving \emph{expensive} optimization problems to generate each \cf{}, and they generate \cf{s} with respect to a \emph{fixed} model, even if solving the same optimization problem for a different model -- such as one that may be adopted in the future -- would yield a better and \emph{still valid} \cf{}. 

Our approach, \technique{}, fundamentally reframes the problem of generating robust \cf{s} in two ways: first, we consider the \emph{multiplicity set} of similar models, noting that this set likely contains models that may be adopted in the future when accounting for slight data shifts. Second, we adapt existing work, CounterNet, which considers the model training and \cf{} generation processes as a single pipeline~\citep{counternet}. Specifically, \technique{} uses the existing conception of training a model that jointly performs classification and \cf{} generation, but promotes \cf{} robustness by optimizing \cf{} generation over a multiplicity set of classifiers, rather than a single classifier.  
A key part of how \technique{} obtains robustness is to use verified training~\citep{ibp,crownibp} to \emph{deterministically certify} when the predictions and the \cf{s} our model generates are robust to small model changes, as indicated by an $l_p$-bound on the classifier's parameters, subject to restrictions that hold the original prediction constant.
We develop a new variation of verified training, \ourabsname, that allows us to obtain tighter bounds than existing approaches. 
We show that \cf{s} produced by \technique{} are not only certifiably robust to small model shifts, but are also robust to other empirical forms of model updates---e.g., training with different random seeds or different data subsets. An added benefit is that \cf{} generation using \technique{} is very fast (equivalent to inference speed), with only a modest increase in training time.
 
In summary, we make the following key contributions: (1) We propose robust \cf{} generation over a multiplicity set of models, (2) we develop a loss function to jointly train an accurate model and a robust and valid \cf{} generator, (3) we design a new verified training algorithm, \ourabsname, to soundly overapproximate our loss function during training, and (4) we show that our technique, \textbf{Veri}fied \textbf{Tra}ining for \textbf{CE} \textbf{R}obustness (\technique{}), achieves high \cf{} robustness, both to $l_p$-bounded model shifts and real-world model updates.

\section{Related Work}\label{sec:related}

\paragraph{Explainable AI} 
Explanations for model predictions can come from insights about the model (linear regression coefficients, decision tree rules, or feature activations), or more commonly for black-box models, from a post-hoc technique. Post-hoc explanation techniques are typically either feature-based (e.g., \cite{shap,lime,Simonyan2013DeepIC,smilkov2017smoothgrad}) and aim to describe which input features are relevant, or counterfactual-based (e.g., \cite{karimi2020model,looveren2019interpretable,face,ustun2018actionable,wachter2017counterfactual}) and aim to describe how the original instance needs to change to get a different prediction. In this work, we focus on \cf{s} due to the fact that they provide direct guidance to users.
\cf{s} are typically found post-hoc, based on an optimization problem over a fixed model. An exception is the CounterNet training procedure~\citep{counternet}, which jointly trains a model and \cf{} generator.

\paragraph{Robust explanations} 
``Explanation robustness'' refers to multiple phenomena in the literature; the focus can be robustness with respect to the input (i.e., whether perturbing the input yields a similar explanation) or with respect to model changes (i.e., whether changing the model yields a similar prediction and explanation for a fixed input). In this work, we focus on the latter definition. 

Various works have explored robust \cf{} generation for specific model classes, such as for tree-based ensembles~\citep{dutta2022robust,forel2022robust} or (locally) linear models~\citep{bui2022counterfactual, nguyen2023distributionally}.  Other work on more complex model classes has shown that it is possible to adversarially change a model to keep identical predictions, but drastically change the associated feature-based explanation~\citep{anders2020fairwashing,heo2019fooling,slack2020fooling}. Similarly, a given \cf{} can be invalidated by another equivalently-accurate model in a neural network setting~\citep{hamman2023robust}. But it is still possible to improve (average) explanation robustness to model shift, such as by increasing the local smoothness of the model~\citep{black2021consistent,dombrowski2022towards,meyer2023minimizing,srinivas2022efficient}. For \cf{s}, increasing the distance to the \cf{} is also commonly thought to increase robustness to model shift~\citep{drobustness_inn,pawelczyk2020counterfactual}. However, this heuristic does not always work for deep models~\citep{black2021consistent}; instead, the agreement of points in an epsilon-ball around the \cf{} is important~\citep{hamman2023robust}. However, all of these papers consider \cf{} generation with respect to a fixed model -- by contrast, our process jointly trains a model and \cf{} generator in order to obtain robustness. Also, most existing methods only evaluate \cf{} robustness based on empirical model shift. An exception is \cite{drobustness_inn}, who consider \cf{} robustness to a set of models (via $l_p$-norm-bounded changes to the parameters of a fixed model). We consider the same $l_p$-norm bounded setting, but also evaluate our approach on empirical model shifts.





\section{Problem Definition}
\label{sec:prob_def}
Let $f$ be a neural net classifier that learns its parameters $\param{f}$
with a loss function $\loss$ and a training set $D=\{(\x_i, y_i)\}_{i=1}^n$. 
We will assume that $f$ is a binary classifier, i.e., assigns $\x$ a label $y\in \{0, 1\}$. 
A distribution shift may occur and yield a new training set $D'=\{\x_i,y_i)\}_{i=1}^k$. 
We \emph{finetune} $f$ (i.e., train for an additional small number of epochs) or retrain from scratch in the presence of $\ell_2$ regularization to yield a shifted model $\rs$. Given $f$ and $\rs$ with parameters $\param{f}$ and $\param{\rs}$, respectively, we measure their distance $\|\param{f}-\param{\rs}\|_p $ as the largest $\ell_p$ parameter distance among all layers (both weight matrices and bias vectors) in $f$ and $\rs$. We make the assumption that the finetuning (or retraining with regularization) process will yield a model $\rs$ such that $\|\param{f}-\param{\rs}\|_p\leq \delta$ for some small $\delta$. 
For simplicity, we use a single $\delta$ for the $l_p$ bound across all layers. However, our approach easily extends to using different $\delta$ values for different layers.


\paragraph{Counterfactual explanations} A counterfactual explanation (\cf{}) generator $\gen$ returns a \cf{} $\x'$ with respect to a function $f$ and an original input $\x$, i.e., $\x'=\gen(f,\x)$. We say that $\x'$ is \emph{valid} if it satisfies $f(\x')\neq f(\x)$. The distance between $\x$ and $\x'$ is called \emph{proximity} and should be minimized subject to the validity constraint. Additional constraints (e.g., not changing immutable features like race or gender) can be placed on the \cf{} generation process, e.g., by using a custom distance metric that heavily penalizes changing those features. Our approach is compatible with that type of modification, but we assume an $l_1$-norm distance metric for simplicity.


\paragraph{Multiplicity set of $f$} Even though $f$ is the result of an optimization process, it likely is not the only model that performs well on the training data due to model multiplicity~\citep{damourUnderspecification2022,marx2020predictive}. Furthermore, the finetuning or retraining process will yield a distinct model $\rs$ that is close to $f$.  
We adopt a common assumption from the literature~\citep{drobustness_inn,roar} that the parameter distance between $f$ and $\rs$ will be bounded by $l_p$ norms. We define the \emph{multiplicity set} with respect to a model $f$ and an input $\x$ as follows.

\begin{definition}[Multiplicity set]
\label{def: multiplicityset}
Given a model $f$ with parameters $\param{f}$, an input $\x$, an $l_p$ norm, and a bound $\delta$, we define the \textit{$\delta$-robust multiplicity set} 
as
$\mult_{f,\x} = \{ \rs \mid f(\x) = \rs(\x) \wedge \|\param{f} - \param{\rs}\|_{p} \le \delta \}$.
\end{definition}
Intuitively, the multiplicity set contains models that have comparable performance to $f$ and have a similar set of model weights. Crucially, it also is intended to contain models $f_m$ that may be the result of model updates due to distribution shift. Note that our definition limits us to models with the same prediction on $\x$. As we are primarily interested in \cf{s}, it no longer makes sense to consider adopting the \cf{} if the prediction changes. 

\paragraph{\cf{} robustness}
\emph{Data shift necessitates that models change over time}, thus creating the risk that \cf{s} generated by a model will be invalidated in the future.
We define \emph{\cf{} robustness} as follows.

\begin{definition}[$\mult_{f,\x}$-robustness of a \cf{}]
    Given a binary classifier $f$ and an input $\x$, we say that a \cf{} $\x'$ is \emph{robust} if it is a valid \cf{} for all models in the multiplicity set, i.e., if $\forall \rs \in\mult_{f,\x}$, $\rs(\x') = 1 - f(\x)$.
\end{definition}

\paragraph{Goal}
Our goal is to devise a training algorithm that yields a model $f$ and a \cf{} generator $\gen$ such that the \cf{s} generated by $\gen$ will be $\mult_{f,\x}$-robust (i.e., robust to small changes in the model $f$).
In other words, we want to maximize the number of robust \cf{s} that we generate on some dataset (e.g., on the training dataset during training, or on a test or validation dataset post-training). 
Note that our goal differs from all existing works, which propose algorithms to generates robust \cf{s} according to a fixed model $f_\mathrm{fix}$.
The following example illustrates how focusing on the model selection process---rather than just on the \cf{} generation technique---can yield better and more robust \cf{s}. 

\begin{figure*}[t]
    \centering
    \begin{subfigure}[b]{0.3137\textwidth}
        \includegraphics[width=\textwidth]{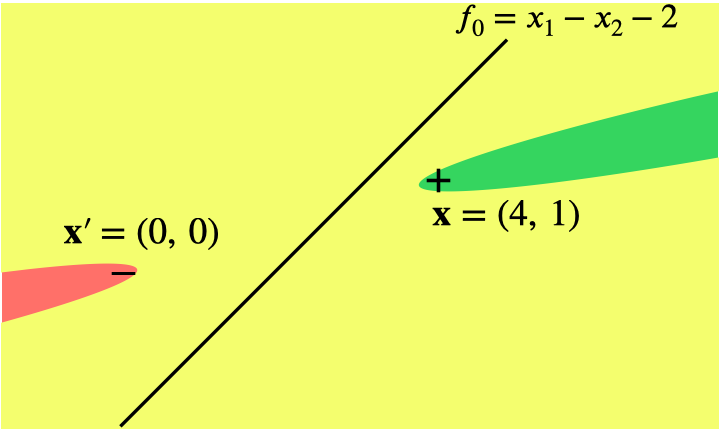}
        \caption{\small{$f_0=x_1-x_2-2$. The optimal robust \cf{} of $\x=(4,1)$ is $(0,0)$.}}
    \end{subfigure}
    \hfill
    \begin{subfigure}[b]{0.3137\textwidth}
        \includegraphics[width=\textwidth]{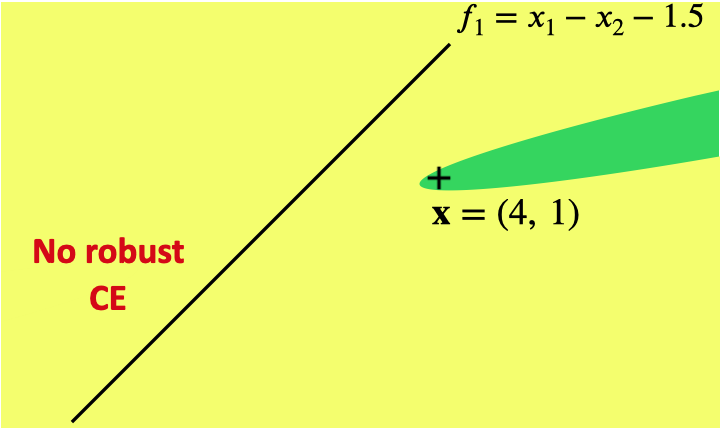}
        \caption{\small{$f_1=x_1-x_2-1.5$. There exists no robust \cf{} of $\x=(4,1)$.}}
    \end{subfigure}
    \hfill
    \begin{subfigure}[b]{0.3137\textwidth}
        \includegraphics[width=\textwidth]{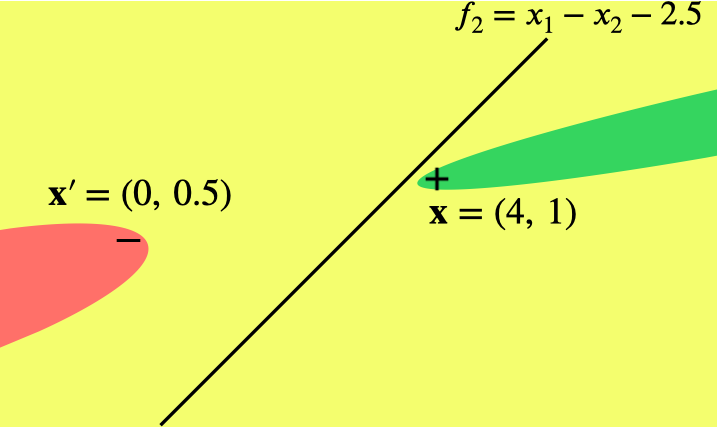}
        \caption{\small{$f_2=x_1-x_2-2.5$. The opt. robust \cf{} of $\x=(4,1)$ is $(0,0.5)$.}}
    \end{subfigure}
    \vspace{-0.5em}
    \caption{\small{Plots of three linear models and their multiplicity sets. 
    The black line shows the original linear model $f_i$.
    The green and red regions contain all samples that are robust under the multiplicity set $\mult_{f_i, \x}$, receiving predictions $+$ and $-$, respectively. The yellow region corresponds to all samples that are not robust under the multiplicity set $\mult_{f_i,\x}$.}}
    \label{fig:example1_plots}
    \vspace{-1em}
\end{figure*}

\begin{example} 
    Suppose we have an input $\x=(4,1)$ and the three equally-accurate linear models $f_0,f_1,f_2$ shown in \cref{fig:example1_plots}. Considering that these models are equally accurate, we can use \cf{} robustness to small model shifts as a secondary criteria for choosing the best model. To do so, 
    we consider the multiplicity set $\mult_{f_i, \x}$ of $f_i$ containing linear models with a bound $\delta=2$ on the $l_\infty$ norm. In \cref{fig:example1_plots}, the yellow regions represent all samples that are not robust to $\mult_{f_i,\x}$.
    %
    The optimal robust \cf{} of $\x$ on $f_0$ is $\cfx{\x}_0 = (0,0)$. 
    However, $\x$ has no robust \cf{} on $f_1$, and its optimal robust \cf{} on $f_2$ is $\cfx{\x}_2 = (0, 0.5)$. In other words, if $f_\mathrm{fix}=f_1$, then no \cf{} generation algorithm $\gen$ will be able to find a robust counterfactual. But if $f_\mathrm{fix}=f_2$, rather than $f_0$, we can find a better robust \cf{} (since $\cfx{\x}_2$ is closer to $\x$ than $\cfx{\x}_0$ is). So, if we train $f$ and $\gen$ jointly, we can try to end up with a model closer to $f_2$: that is, a model that performs well and also yields robust \cf{s} that are of high-quality (i.e., close to the original sample).
\end{example}

\section{Approach}
\label{sec:approach}
Our approach, \technique{}, is a training algorithm that takes a training set $D$ as input and outputs a model $f$ and a \cf{} generator $\gen$ such that the \cf{s} generated by $\gen$ have a high rate of $\mult_{f,\x}$-robustness. 
%
Devising such a training algorithm requires us to solve the following two challenges.
First, training a model to produce robust explanations requires reasoning in tandem about the accuracy of the model $f$ and the validity, quality, and robustness of the \cf{s} produced by the \cf{} generator $\gen$ (\cref{sec:tandem_loss}).  
Second, we need to be able to \emph{soundly overapproximate} our robust loss function, as using gradient descent to approximate the loss will be both inefficient and may overlook some $\rs\in\mult_{f,\x}$ (\cref{sec:overapprox_loss}).

\subsection{Training the model $f$ and the \cf{} generator $\gen$ in tandem}\label{sec:tandem_loss}
To intertwine model training and robust \cf{} generation, we build off of CounterNet~\citep{counternet}, which jointly learns a model $f$ and a (non-robust) \cf{} generator $\gen$. Intuitively, we have two loss functions, $\lossf$ and $\lossgen$, that are used to optimize $f$ and $\gen$, respectively. In each training epoch, there are three steps. First, we generate \cf{s} $\cfx{\x}_i = \gen(f,\x_i)$ according to the current weights of $f$ and $\gen$. Next, we optimize the parameters of $f$ using gradient descent (WRT $f$) on $\lossf$, and finally, we optimize the parameters of $\gen$ using gradient descent (WRT $g$) on $\lossgen$.  

We denote the output of model $f$ after the final sigmoid layer but before discretization as $\spred{f}$. The function $\spred{f}(\x)\in[0,1]$ is frequently employed in loss functions for improved optimization compared to the hard label $f(\x)\in\{0,1\}$.
We define helper loss functions $\lossacc$, $\lossval$, and $\lossqual$ to promote accuracy, \cf{} validity, and \cf{} quality (i.e., proximity to the original sample $\x_i$), respectively. All three losses were used in the original CounterNet loss, and can be any binary loss function, e.g., MSE. $\lossrobust$ is our proposed \robustcfx loss, and aims to measure the robustness of a \cf{} $\cfx{\x}$ with respect to $\mult_{f,\x}$ (as defined by hyperparameters $\delta$ and $p$, which are omitted in the equations below for brevity). This loss function should capture the worst classifier $\rs\in\mult_{f,\x}$, i.e., the model that does most poorly on the \cf{} $\cfx{\x}$. We define $\lossrobust(\x,\cfx{\x},\param{f})=\max_{\rs\in\mult_{f,\x}} \lossmse(\spred{\rs}(\cfx{\x}), 1-y)$. 
With those definitions in mind, we define $\lossf$ and $\lossgen$ as follows:

\vspace{-1em}
\begin{small}
\begin{equation}
    \lossf (\x_i, y_i, \cfx{\x}_i, \param{f}) = \lambda_1 \lossacc(\spred{f}(\x_i), y_i) + \lambda_2 \lossrobust(\x_i,\cfx{\x}_i,\param{f})
\end{equation}
\begin{equation}
     \lossgen(\x_i,y_i,\gen(f,\x_i), \param{f}) = \lambda_3 \lossqual(\x_i,\gen(f,\x_i)) + \lambda_4 \lossval(\spred{f}(\gen(f,\x_i)), 1-y_i) + \lambda_2 \lossrobust(\x_i,\gen(f,\x_i), \param{f})
\end{equation}
\end{small}


\subsection{Computing the \robustcfx loss}\label{sec:overapprox_loss}

Exactly minimizing $\lossrobust$ is expensive: we either need to consider infinitely many reasonable models in $\mult_{f,\x}$ or we need to rely on expensive gradient descent to approximate the model $\rs$ with no guarantee that it is the worst-case model.
To address this, we use abstract interpretation~\citep{DBLP:conf/popl/CousotC77} 
to efficiently compute an upper bound $\abs{\lossrobust}(\x,\cfx{\x},\param{f})$ 
on the \robustcfx loss. 
This upper bound, when minimized and sufficiently tight, ensures a reduction in the \robustcfx loss simultaneously.
The challenge of overapproximation is how to compute the upper bound $\abs{\lossrobust}(\x, \cfx{\x}, \param{f})$ as tightly as possible: overly loose bounds may hamper prediction accuracy, \cf{} validity, and \cf{} quality, without meaningfully enhancing \cf{} robustness.

In \cref{sec:ibp}, we show how to compute  $\abs{\lossrobust}$ using two existing abstract interpretation techniques, IBP~\citep{ibp} and CROWN-IBP~\citep{crownibp}. Then, in \cref{sec: ourapproach} we introduce a new technique that yields tighter upper bounds on the robust loss while maintaining efficiency similar to CROWN-IBP. 

\subsubsection{Overapproximating the \robustcfx loss using existing techniques} \label{sec:ibp}

Interval bound propagation (IBP) allows us to evaluate a function on an infinite set of inputs represented as a hyperrectangle in $\mathbb{R}^n$.
We will use an interval $\interval{\generalzb} = [\lb{\generalzb},\ub{\generalzb}]$, where $\lb{\generalzb}, \ub{\generalzb} \in \mathbb{R}^n$ and $\forall 1\le i \le n,\ \lb{\generalzb}_i \le \ub{\generalzb}_i$, to denote the set of all $n$-dimensional vectors whose $i$-th element is between $\lb{\generalzb}_i$ and $\ub{\generalzb}_i$, inclusive. 
Previous work~\citep{ibp,arc} has used IBP to overapproximate the worst-case loss of test-time adversarial attacks. 
These works look at perturbations of an input $\x$, i.e., an infinite set of possible $\x$, as evaluated on a fixed model. 
By contrast, we want to use IBP to overapproximate the infinitely many models in $\mult_{f,\x}$.  

To use IBP, we will relax the definition of $\mult_{f,\x}$ to $\mult_{f}$ by removing the requirement that the classification of $\x$ remains constant, i.e., $\mult_{f} = \{\rs \mid \|\param{f} - \param{{\rs}}\|_p \le \delta\}$.
After relaxation, the components of $\param{f}$ (the weights and biases in linear layers) can be overapproximated by intervals.
Therefore, we apply IBP to overapproximate the \robustcfx loss under the multiplicity $\mult_{f}$ using interval arithmetic.
\begin{wrapfigure}{r}{0.4\textwidth} 
  \centering
  \vspace{-1em}
  \includegraphics[width=0.38\textwidth]{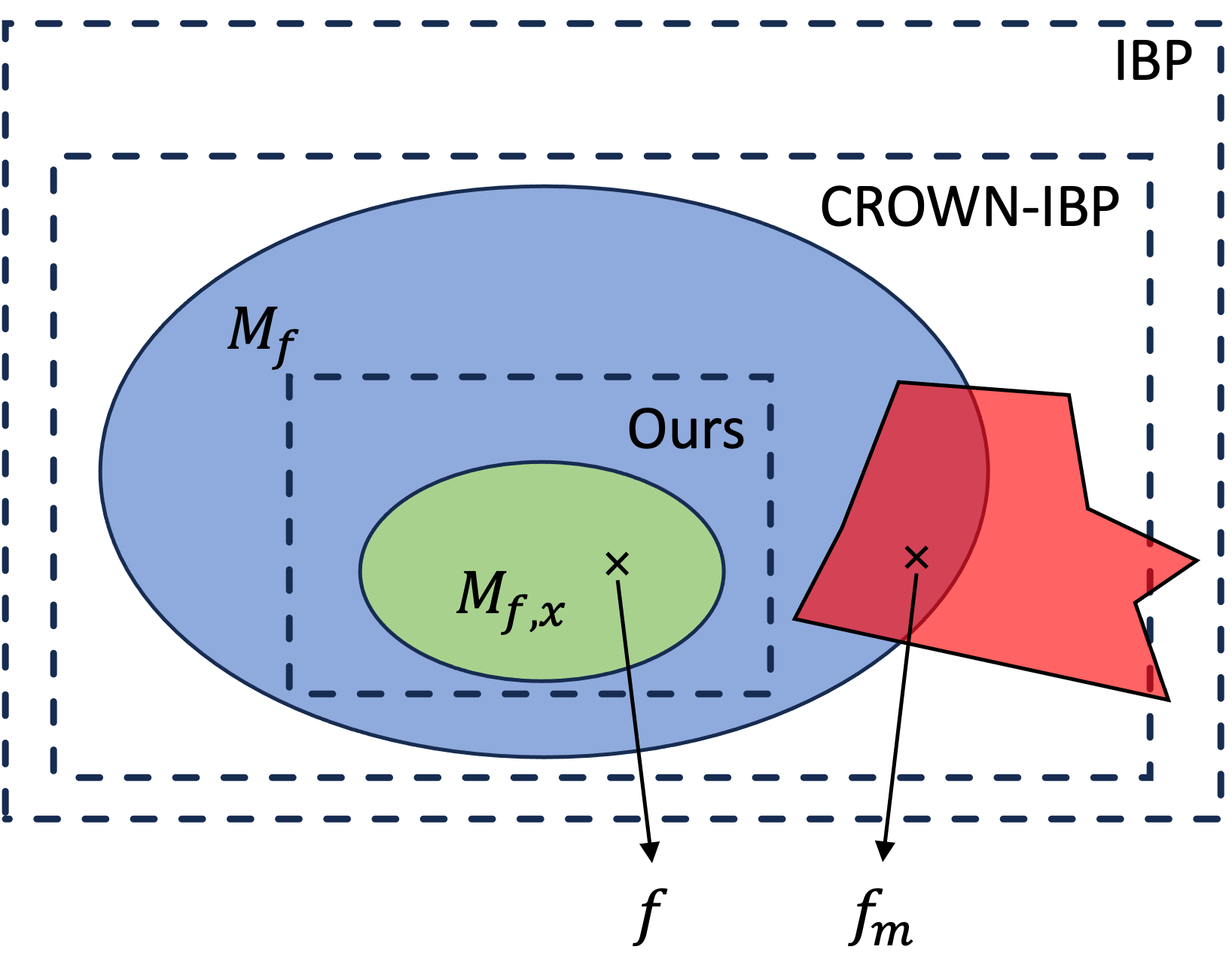} 
  \vspace{-0.5em}
  \caption{Our approach \ourabsname achieves a tighter overapproximation than IBP and CROWN-IBP because the latter techniques include portions of the red region where the \cf{} is not robust.}
  \vspace{-1em}
  \label{fig: illustrate}
\end{wrapfigure}


CROWN-IBP~\citep{crownibp} achieves tighter bounds than IBP by incorporating a backward tightening technique employed in $\alpha$-CROWN~\citep{acrown, deeppoly}.
In the evaluation of a function $f_{\interval{\param{}}}$  whose 
parameters $\interval{\param{}}$ are intervals, CROWN-IBP finds
$\lbalpha,\ubalpha,\lbbeta, \ubbeta$  such that 
$\sigma(\lbalpha \param{} + \lbbeta) \leq \spred{f}_{\param{}}(\x) \leq \sigma(\ubalpha \param{} + \ubbeta)$ for all $\param{}\in\interval{\param{}}$. 
Then, these equations ($\lbalpha \param{} + \lbbeta$ and $\ubalpha \param{} + \ubbeta$) serve to bound $\spred{f}$, allowing us to do IBP with tighter -- yet still sound -- bounds. 
%
%
Similar to the adaption of IBP, we need to relax the definition of $\mult_{f,\x}$ to $\mult_{f}$ for CROWN-IBP because it cannot deal with the additional requirement of prediction robustness that $\mult_{f,\x}$ requires.
After this relaxation, the parameters in all layers (i.e., the weight matrices and bias vectors),  can be overapproximated into intervals.
When applying CROWN-IBP to overapproximate the \robustcfx loss, the interval $\interval{\bfz}$ becomes the intervals of parameters in all layers, and the $\lbalpha, \ubalpha, \lbbeta, \ubbeta$ are computed based on a specific input $\x$ and the lower and upper bounds of parameters, i.e., $\lb{\bfz}$ and $\ub{\bfz}$.



\begin{theorem}[Soundness and Tightness]
\label{theorem: cibp}
For any $\x$, $\cfx{\x}$, and $f$, the CROWN-IBP-overapproximated loss $\abscibp{\lossrobust}(\x, \cfx{\x}, \param{f})$ is an upper bound of the \robustcfx loss $\lossrobust(\x, \cfx{\x}, \param{f})$ and a lower bound of the IBP-overapproximated loss $\absibp{\lossrobust}(\x, \cfx{\x}, \param{f})$. Formally,
\begin{small}
    \[\absibp{\lossrobust}(\x, \cfx{\x}, \param{f}) \ge \abscibp{\lossrobust}(\x, \cfx{\x}, \param{f}) \ge \lossrobust(\x, \cfx{\x}, \param{f})\]
\end{small}
\end{theorem}

\begin{example}
    \label{ex:cibp}
    Consider an interval linear layer with weight matrix $\interval{W} = ( [-1,3], [-3,1])$ and bias scalar $\interval{b}=[-4,0]$. For an input $\x=(-4,-1)$, CROWN-IBP concatenates $\interval{W}$ and $\interval{b}$ to form $\interval{\param{}} = ([-1,3], [-3,1],[-4,0])$ and computes $\lbalpha=\ubalpha=(\bfx;1)=(-4,1,1)$ (note that we append 1 to $\x$ as the multiplicative factor for $\interval{b}$) and $\lbbeta=\ubbeta=0$. Then we can symbolically compute the output based on $\lbalpha{}^T\param{} + \lbbeta$ (or $\ubalpha{}^T\param{} + \ubbeta$ because the two are the same), i.e., $(-4,-1,1)^\top([-1,3], [-3,1], [-4,0])$.
    To compute the concrete upper bounds, we apply interval arithmetic to the vector product and get $[-17,7]$.
    Note that IBP and CROWN-IBP yield identical lower and upper bounds for this linear model because it has only one layer.
\end{example}

However, the bounds computed by CROWN-IBP are still loose due to the relaxation from $\mult_{f,\x}$ to $\mult_{f}$ as illustrated in the following example.

\begin{example}
\label{ex:ibp_cbip_loose}
Consider the same interval linear model as in \cref{ex:cibp}. Note that the midpoint of this interval network corresponds to our ``true'' $f$, i.e., $W=(1,-1)$ and $b=-2$. Let $\spred{f}(\x)=\sigma(W\x+b)$. For the input $\x=(4,1)$ and its \cf{} $\cfx{\x}=(-4,-1)$, we have $\spred{f}(\x)>0.5$ and $\spred{f}(\cfx{\x})<0.5$. 
IBP and CROWN-IBP are unable to prove that the \cf{} $\cfx{\x}$ is robust even though $\mult_{f, \x}$ does not intersect with the ``unsafe'' red region shown in Figure~\ref{fig: illustrate}. 
To see this limitation, let us consider a model $\rs$ with $W'=(-1,-3)\in \interval{W}$ and $b'=0\in\interval{b}$ that is in $\mult_{f}$ but outside $\mult_{f, \x}$ (the output of $\spred{\rs}(\x)$ is $\sigma(-7) <0.5$).
$\rs$ has an output of $\sigma(W'\cfx{\x} + b')=\sigma(7)>0.5$ on the \cf{} $\cfx{\x}$ meaning that the \cf{} is invalid on $\rs$. 
%
%
%
This limitation arises from the relaxation from $\mult_{f,\x}$ to $\mult_{f}$. 
Our approach (\cref{sec: ourapproach}) overcomes this limitation by overapproximating $\mult_{f,\x}$ instead of its relaxation $\mult_{f}$ and is able to prove that $\cfx{\x}$ is $\mult_{f, \x}$-robust.
\end{example}

\subsubsection{Overapproximation the \robustcfx Loss using \ourabsname} \label{sec: ourapproach}

The primary challenge faced by IBP and CROWN-IBP in handling $\mult_{f,\x}$ is their inability to \emph{simultaneously} reason about the overapproximation of $f(\x)$ and $f(\cfx{\x})$.
In this section, we present our approach, \ourabsname, which addresses this challenge to achieve a more precise overapproximation of the \robustcfx loss.

To simultaneously overapproximate $f(\x)$ and $f(\cfx{\x})$, we first reframe the definition of the \robustcfx loss, $\lossrobust = \max_{\rs\in\mult_{f,\x}} \lossmse(\spred{\rs}(\cfx{\x}), 1-\hat{y})$. Namely, if $\hat{y}=1$, $\lossrobust$ is maximized when the worst-case classifier is $[ \mathrm{arg}\max_{\rs \in \mult_{f}} \spred{\rs}(\cfx{\x}) \quad \textrm{s.t. } \spred{\rs}(\x) \ge 0]$. 
However, if $\hat{y}=0$, $\lossrobust$ is maximized when the worst-case classifier is $[\mathrm{arg}\max_{\rs \in \mult_{f}} \spred{\rs}(\cfx{\x}) \quad \textrm{s.t. } \spred{\rs}(\x) \ge 0]$. 
%
Using this insight, we apply CROWN-IBP to obtain lower and upper bounds in $\mult_{f}$ for each training instance $\x_i$.
For all parameters $\param{} \in \interval{\param{}}$,
we have $\sigma(\lbalpha_i \param{} + \lbbeta_i) \leq \spred{f}_{\param{}}(\x) \leq \sigma(\ubalpha_i \param{} + \ubbeta_i)$ 
and $\sigma(\lbmu_i \param{} + \lbnu_i) \leq \spred{f}_{\param{}}(\x) \leq \sigma(\ubmu_i\param{} + \ubnu_i)$. 
We compute $\boldsymbol{\alpha}_i$ and $\beta_i$ based on $\x_i$ and $\interval{\param{}}$, while $\boldsymbol{\mu}_i$ and $\nu_i$ are computed based on $\cfx{\x}_i$ and $\interval{\param{}}$. 


\begin{theorem}[Overapproximation by \ourabsname]
    Conditional on the value of $\hat{y}$, the optimal value of $\lossrobust$ can be upper bounded by the solution to one of the following cases,
    \begin{align}
   \begin{cases}
      t_{\hat{y}=1} =  \underset{\param{} \in \interval{\param{}}}{\max}~~\ubmu_i \param{} + \ubnu_i \quad \textrm{s.t. } \ubalpha_i \param{} + \ubbeta_i \geq 0, &
      \\
     t_{\hat{y}=0} =   \underset{\param{} \in \interval{\param{}}}{\min}~~~\lbmu_i \param{} + \lbnu_i ~\quad \textrm{s.t. } \lbalpha_i \param{} + \lbbeta_i \leq 0, &
     \label{eq:sim_overapprox}
   \end{cases}
    \end{align} 
    That is, if $\hat{y}=1$, we have $\lossrobust \leq \lossmse(\sigma(t_{\hat{y}=1}), 0)$, and if $\hat{y}=0$, $\lossrobust\leq \lossmse(\sigma(t_{\hat{y}=0}), 1)$.
\end{theorem}

We can use a solver to address each case in  \Cref{eq:sim_overapprox} by encoding it into a linear programming (LP) problem. However, solving an LP for each training batch is time-consuming and breaks the gradient information required to optimize the \robustcfx loss.
Appendix~\ref{appendix: algo} introduces a greedy algorithm (\cref{alg: greedy_solve}) that solves \cref{eq:sim_overapprox} in $O(n\log n)$ time, where $n$ is the size of parameters, and can be implemented in PyTorch preserving the gradient information.

\begin{theorem}[Soundness and Tightness]
\label{theorem: ours}
For any $\x$, $\cfx{\x}$, and $f$, the overapproximated loss $\absours{\lossrobust}(\x, \cfx{\x}, \param{f})$ is an upper bound of the \robustcfx loss $\lossrobust(\x, \cfx{\x}, \param{f})$ and a lower bound of the CROWN-IBP-overapproximated loss $\abscibp{\lossrobust}(\x, \cfx{\x}, \param{f})$. Formally,
\begin{small}
    \[\abscibp{\lossrobust}(\x, \cfx{\x}, \param{f}) \ge \absours{\lossrobust}(\x, \cfx{\x}, \param{f}) \ge \lossrobust(\x, \cfx{\x}, \param{f})\]
\end{small}
\end{theorem}
Note that $\absours{\lossrobust}(\x, \cfx{\x}, \param{f}) = \abscibp{\lossrobust}(\x, \cfx{\x}, \param{f})$ when for all $1 \le i \le n$, $\ubs{\cfxslope}_i\ubs{\slope}_i > 0$ if $\hat{y}=1$ (or $\lbs{\cfxslope}_i\lbs{\slope}_i > 0$ if $\hat{y}=0$). 

The following example shows that \ourabsname is tighter than the CROWN-IBP-overapproximated loss.

\begin{example}
    \label{ex: our_tight}
    Consider the same linear model given in \cref{ex:ibp_cbip_loose}, for the original input $\x=(4,1)$ and the corresponding \cf{} $\cfx{\x}=(-4,-1)$, we have the following optimization problem according to \cref{eq:sim_overapprox}.
    \begin{align*}
        \underset{\bfz \in \interval{\bfz}}{\max} -4z_1-z_2+z_3  \quad \textrm{s.t. } 4z_1+z_2+z_3 \ge 0,
        \text{ where } \lb{\bfz} = [-1, -3, -4], \ub{\bfz} = [3, 1, 0]
    \end{align*}
    According to \cref{alg: greedy_solve}, the optimal value $0$ is obtained when $\bfz=[-\frac{1}{4},1,0]$. In other words, the output of the worst-case classifier in $\mult_{f,\x}$ is $\sigma(0)$ computed by \ourabsname. This upper bound is tighter than the ones computed by IBP and CROWN-IBP ($\sigma(7)$ in \cref{ex:ibp_cbip_loose}).
\end{example}

\section{Experimental Evaluation}\label{sec:eval}

In this section, we present our evaluation of \technique{}. First, we describe the experiment setup (\cref{sec:ex-setup}), and next we show results (\cref{sec:ex-results}) on \deltarobust, \nfoldrobust, and \cf{} quality. 

\subsection{Experimental setup} \label{sec:ex-setup}
\paragraph{Datasets}
We perform our evaluation on two \emph{real-world distribution shift} datasets, Cardiotocography (CTG)~\citep{cardiotocography} and WHO~\citep{whoDataset}, as well as three other datasets commonly used in the robust counterfactual literature, Home Equity Line of Credit (HELOC)~\citep{heloc}, Taiwanese Credit (TC)~\citep{taiwanCredit}, and Open University Learning Analytics (\student)~\citep{kuzilek2017open}. CTG aims to predict whether a fetal cardiotocogram is healthy, WHO aims to predict whether a country's life expectancy is above the median, HELOC aims to predict the likelihood that home buyers will repay a loan within 2 years, TC aims to predict default on their credit card payments, and \student aims to predict whether a student will pass an online class. Additional details including dataset composition and preprocessing steps such as binarization and distribution shift construction are available in the appendix.

\paragraph{Metrics}
Given an $\x$ and a corresponding \cf{} $\cfx{\x}$, \deltarobust is satisfied when we can certify that $\cfx{\x}$ obeys $\rs(\x')=f(\cfx{\x})$ for all $\rs\in\mult_{f,\x}=\{\rs \mid f(\x) = \rs(\x) \land \|\param{\rs}-\param{f}\|_p \leq \delta\}$ using \ourabsname. We dynamically determine distinct $\delta_i$ values for each layer (see the appendix for details); for simplicity, we will continue to write \deltarobust as if $\delta$ is a constant. 
By contrast, $\cfx{\x}$ exhibits \nfoldrobust with respect to two models $f$ and $\rs$ when $f(\cfx{\x})=\rs(\cfx{\x})$. In practice, we compute empirical \nfoldrobust
 across sets of models trained with one of three variations: different random initialization (RI) for training $f$ and $\rs$, different training datasets constructed by randomly removing 1\% of the data (LOO) for $f$ and $\rs$, or different training datasets under data shift (DS) for $f$ and $\rs$. 
We refer to the \emph{\deltarobust rate} as the fraction of data points in a test set that exhibit \deltarobust, and similarly for \emph{\nfoldrobust rate}. 
As a secondary metric, we consider how feasible the counterfactual is to implement through its proximity, i.e., the $l_1$ distance between $\x$ and $\cfx{\x}$. We consider two other distance metrics in the appendix. 

\paragraph{Baseline comparisons}
We compare our approach with the following existing robust counterfactual generation methods: Counternet (CN)~\citep{counternet} without our robustifying modifications; ROAR~\citep{roar}, which finds robust counterfactuals by using adversarial training in the counterfactual generation process; and SNS~\citep{black2021consistent}, which finds \cf{s} in regions with a low Lipschitz constant. For ROAR and SNS, we train NNs with the same architecture as the predictor part of CounterNet; we optimize this model using a loss function that solely prioritizes accuracy. We implement all three techniques (IBP, CROWN-IBP, \ourabsname) from \cref{sec:overapprox_loss}. We include results for \ourabsname here and put results for the others in the ablation study in the appendix.

\paragraph{Experimental procedure}
To evaluate \deltarobust, we select layer-specific $\delta_i$ (see appendix for details) and perform training and all evaluation for that setting. 
To evaluate \nfoldrobust for RI and LOO, we train 10 models, generate \cf{s} for each, and report the average fraction of these \cf{s} that remain valid across the other 9 models.  To evaluate \nfoldrobust for DS, we train a model on the original data, then finetune for a small number (typically 20) additional epochs on the new data to obtain the shifted model.


\begin{table}[t]
    \small
    \centering
    \caption{\small Fraction of samples that are pair-wise \nfoldrobust across 10 model trained with different random initializations (RI) or different segments of data randomly removed (leave-one-out, or LOO). Standard deviations are in parenthesis. Best result is in \textbf{bold} and second-best result is \underline{underlined}.}
    \begin{tabular}{l|rrr|rrr}\toprule
    & \multicolumn{3}{c|}{Random Initialization} & \multicolumn{3}{c}{Leave-One-Out}\\
   & \multicolumn{1}{c}{HELOC} & \multicolumn{1}{c}{TC} & \multicolumn{1}{c|}{OULA} & \multicolumn{1}{c}{HELOC} & \multicolumn{1}{c}{TC} & \multicolumn{1}{c}{OULA} \\\midrule
  \technique{}  & \textbf{0.93} (0.03) & \underline{0.88} (0.14) & \textbf{0.94} (0.05) & 0.86 (0.14) & \underline{0.97} (0.01) & \underline{0.96} (0.05) \\
 CN & \underline{0.92} (0.07) & 0.82 (0.26) &  0.92 (0.07) & \underline{0.91} (0.07) & 0.96 (0.02) &\textbf{0.97} (0.01) \\
 ROAR & 0.88 (0.02) & 0.44 (0.25) &  \underline{0.93}  (0.05)  & 0.88 (0.02) &  0.88 (0.01) & 0.95 (0.04) \\
 SNS & 0.90 (0.04) &\textbf{0.98} (0.02) & 0.80 (0.11)  & \textbf{0.94} (0.05) &  \textbf{1.00} (0.00) & 0.90 (0.06)
      \\\bottomrule
    \end{tabular}
    \label{tab:nfoldrobust:ri}
\end{table} 

\begin{wraptable}{r}{6cm}
\small
    \centering
    \caption{\small Fraction of samples whose \cf{s} are valid and robust after finetuning with a distribution shift. Standard deviation over 10 trials in parentheses.}
    
    \begin{tabular}{l|rr}\toprule
   & CTG  & WHO \\\midrule
 \technique{} & \textbf{0.987} (0.010) & \textbf{0.995} (0.027) \\
 CN & \underline{0.981} (0.019) & \underline{0.967} (0.030) \\
 ROAR & 0.570 (0.104) & 0.884 (0.078) \\
 SNS & 0.407 (0.024) & 0.846 (0.050)
      \\\bottomrule
    \end{tabular}
    \label{tab:distrshift}
\end{wraptable} 

\subsection{Results}\label{sec:ex-results}
\paragraph{\deltarobust}
\technique{} exhibits high levels of \deltarobust: 70.24\% (\textpm 17.41\%) of test samples exhibit \deltarobust for HELOC, 81.08\% (\textpm 4.51\%) for TC, and 96.96\% (\textpm 0.96\%) for OULA. For the same value of $\delta$, CN has a \deltarobust for 26.29\% (\textpm 18.29\%) of HELOC samples, but 0\% for TC and OULA samples. Likewise, the \cf{s} generated by ROAR and SNS for models trained in a standard way have 0\% \deltarobust on all three datasets. 


\paragraph{Random initialization and leave-one-out results}

\Cref{tab:nfoldrobust:ri} shows what fraction of counterfactuals, generated for a particular model, remain valid for a model trained (a) using a different random seed, or (b) with a different 1\% of the training data removed. We 
note that most \cf{} and dataset combinations have high ($>90$\%) \nfoldrobust rates; however, \technique{} has the highest or second-highest \nfoldrobust rate for most settings.

\paragraph{Real-world distribution shifts}
\Cref{tab:distrshift} shows the robustness of \cf{s} to real-world distribution shifts. We see that \technique{} significantly outperforms the non-CounterNet baseline methods, especially on CTG.

\paragraph{Counterfactual Quality}
\begin{wraptable}{r}{8.4cm}
\small
    \centering
    \caption{\small Average proximity for valid \cf{s}. (*) indicates that the \cf{} technique does not correctly account for categorical features.}
    
    \begin{tabular}{l|rrrrr}\toprule
   & CTG  & WHO & HELOC & TC & OULA \\\midrule
 \technique{} & 0.250 & 0.216 & 0.099 & 0.244 & 0.179 \\
 CN & 0.220 & 0.164 & 0.107 & 0.234 & 0.169\\
 ROAR & \underline{0.189} & \underline{0.031} & \textbf{0.032} & \underline{0.018}* & \underline{0.015}* \\
 SNS & \textbf{0.038} & \textbf{0.031} & \underline{0.041} & \textbf{0.015}* & \textbf{0.013}*
      \\\bottomrule
    \end{tabular}
    \label{tab:proximity}
\end{wraptable} 
A weakness of \technique{} is that the generated \cf{s} are a larger distance from the original \cf{s}. The data on proximity (the $l_1$-distance between an input $\x$ and its \cf{} $\cfx{\x}$ is) is summarized in \cref{tab:proximity}. Note that the two datasets where \technique{} performs worst relative to the other techniques (TC and OULA) have categorical features, which ROAR does not modify, and SNS does not handle properly (i.e., by breaking one-hot encodings).

\section{Conclusions}
We have presented \technique{}, a training algorithm that jointly produces a model $f$ and a \cf{} generator $\gen$ such that the \cf{s} generated by $\gen$ will be robust to small changes in the weights of $f$. We do this by minimizing an upper bound on our robust loss, i.e., we minimize the loss on the \emph{worst-case} model in the multiplicity set $\mult_{f,\x}$. We provide a refinement of interval-bound propagation, \ourabsname, allowing the over-approximation to be tighter than other state-of-the-art approaches. Our approach is able to to find \cf{s} that are \emph{certifiably robust} at high rates (typically over 90\%). This carries over to high robustness for empirical model updates, such as retraining with a different random seed. In particular, we outperform state-of-the-art approaches on finetuning after real-world distribution shifts.  The tradeoff is that \technique{} generally yields \cf{s} that are a larger distance form the original sample. In some settings, the additional robustness as well as fast \cf{} generation time may be worth the additional recourse costs, but future work should aim to provide similar deterministic guarantees and empirical performance with smaller \cf{} distances.

\section*{Reproducibility Statement}
Our code is available at \url{https://github.com/ForeverZyh/robust_cfx}.



\bibliography{refs}
\bibliographystyle{iclr2024_conference}

\newpage

\appendix

\section{Algorithms}
\label{appendix: algo}
\Cref{alg:robust_train} gives an overview of our robust training approach. We concurrently train a model $f$ and a \cf{} generator $\gen$. Alternatively, it is possible to use a fixed \cf{} generator and to train $f$ to be a model whose \cf{s} -- with respect to the fixed $\gen$ -- are likely to be robust. In that case, line 1 of the algorithm only initializes $f$, and line 5 of the algorithm is skipped.

\begin{algorithm}[!t]
    \caption{Joint Training of $f$ and $\gen$}
    \label{alg:robust_train}
    \begin{algorithmic}[1]
        \small
        \renewcommand{\algorithmicrequire}{\textbf{Input:}}
        \renewcommand{\algorithmicensure}{\textbf{Output:}}
        \REQUIRE Training set $D={(\x_i, y_i)}_{i=1}^n$, Hyper-parameters $\delta, p$ in \cref{def: multiplicityset}
        \ENSURE Classifier $f$ and robust \cf{} generator $\gen$ \\
        \STATE Initialize $f$ and $\gen$
        \FOR {$e = 1$ to $\mathrm{maxepoch}$}
            \STATE $\cfx{\x} = \gen(f, \x_1), \gen(f, \x_2), \ldots, \gen(f, \x_{n})$ 
            \STATE Optimize $\param{f}$ via $\nabla_{\param{f}} \sum_{i=1}^n \lossf(\x_i, y_i, \cfx{\x}_i, \param{f}, \delta, p)$ 
            \STATE Optimize $\param{\gen}$ via $\nabla_{\param{\gen}} \sum_{i=1}^n \lossgen(\x_i, y_i, \gen(f, \x_i), \param{f}, \delta, p)$ 
        \ENDFOR
        \STATE \textbf{return} $f, \gen$

    \end{algorithmic}
\end{algorithm}



\Cref{alg: greedy_solve} shows our algorithm to solve \cref{eq:sim_overapprox}. Notably, this algorithm -- when implemented in PyTorch -- preserves gradient information. If we were to na\"ively solve a linear programming (LP) problem to optimize the \robustcfx loss within each training batch, the gradient information would be lost and thus the optimization process would not work. 

\begin{algorithm}[!t]
    \caption{Solving \cref{eq:sim_overapprox} when $\hat{y}=1$}
    \label{alg: greedy_solve}
    \begin{algorithmic}[1]
        \small
        \renewcommand{\algorithmicrequire}{\textbf{Input:}}
        \renewcommand{\algorithmicensure}{\textbf{Output:}}
        \REQUIRE Bounded parameters $\interval{\bfz}=(\lb{\bfz},\ub{\bfz})$, and coefficients from CROWN-IBP $\ubs{\cfxslope}, \ubs{\cfxitcpt}, \ubs{\slope}, \ubs{\itcpt}$
        \ENSURE $\underset{\bfz \in \interval{\bfz}}{\max}~~\ubs{\cfxslope} \bfz + \ubs{\cfxitcpt} \quad \textrm{s.t. } \ubs{\slope} \bfz + \ubs{\itcpt} \ge 0$\\
        \FOR {$i = 1$ to $n$}
            \IF {$\ubs{\slope}_i > 0 \vee (\ubs{\slope}_i = 0 \wedge \ubs{\cfxslope}_i > 0)$}
                \STATE $\bfz_i = \ub{\bfz_i}$
            \ELSE
                \STATE$ \bfz_i = \lb{\bfz_i}$
            \ENDIF
        \ENDFOR
        \STATE $s \gets \ubs{\slope}\bfz + \ubs{\itcpt}$, $\cfx{s} \gets \ubs{\cfxslope}\bfz + \ubs{\cfxitcpt}$
        \IF {$s < 0$}
            \STATE \textbf{return} $-\infty$ \COMMENT{Constraint not satisfied}
        \ENDIF
        \STATE $I \gets \{1\le i \le n \mid \ubs{\cfxslope}_i  \ubs{\slope}_i < 0 \}$
        \STATE Sort the indices list $I$ descendingly by $-\frac{\ubs{\cfxslope}_i}{\ubs{\slope}_i}$
        \FOR {$i \in I$}
            \STATE $\delta \gets |\ubs{\slope}_i|(\ub{\bfz} - \lb{\bfz})$, $\cfx{\delta} \gets |\ubs{\cfxslope}_i|(\ub{\bfz} - \lb{\bfz})$
            \IF {$\delta > s$}
                \STATE $\cfx{s} \gets \cfx{s} + \cfx{\delta} \frac{s}{\delta}$
                \STATE \textbf{break}
            \ENDIF
            \STATE $s \gets s - \delta$, $\cfx{s} \gets \cfx{s} + \cfx{\delta}$
        \ENDFOR
        \STATE \textbf{return} $\cfx{s}$
    \end{algorithmic}
\end{algorithm}

\section{Proofs}
We provide proofs of Theorems \ref{theorem: cibp} and \ref{theorem: ours}.

\begin{proof}[Proof of \Cref{theorem: cibp}]
    First, we will show that the IBP loss provides a sound upper bound on $\lossrobust$. 
    By the soundness of IBP, we have 
    \[\absibp{\lossrobust}(\x, \cfx{\x}, \param{f}) \ge \max_{\rs \in \mult_{f}} \lossmse(\spred{\rs}(\cfx{\x}), 1-\hat{y})\]
    Next, note that $\mult_{f,\x} \subseteq \mult_{f}$ ensures the overapproximation can still capture the worst-case classifier. Formally,
    \[\max_{\rs \in \mult_{f}} \lossmse(\spred{\rs}(\cfx{\x}), 1-\hat{y}) \ge \lossrobust(\x, \cfx{\x}, \param{f})\]

    The proof of CROWN-IBP loss is tighter than the IBP loss is given in \citet{crownibp}.
\end{proof}

\begin{proof}[Proof of \Cref{theorem: ours}]
    Without loss of generality, we consider the case $\hat{y}=1$.
\ourabsname applies CROWN-IBP to obtain lower and upper bounds in $\mult_{f}$.
For all parameters $\param{} \in \interval{\param{}}$,
we have $\sigma(\lbalpha \param{} + \lbbeta) \leq \spred{f}_{\param{}}(\x) \leq \sigma(\ubalpha \param{} + \ubbeta)$ 
and $\sigma(\lbmu \param{} + \lbnu) \leq \spred{f}_{\param{}}(\x) \leq \sigma(\ubmu\param{} + \ubnu)$. 
The coefficients $\lbalpha, \lbbeta, \ubalpha, \ubbeta, \lbmu, \lbnu, \ubmu, \ubnu$ in \ourabsname are the same as the ones in CROWN-IBP.

\ourabsname computes the upper bound of $\lossrobust$ as $\absours{\lossrobust}(\x, \cfx{\x}, \param{f}) =\lossmse(\sigma(t_\emph{Ours}), 0)$, where $t_\emph{Ours}$ is obtained in the following optimization problem,
\begin{align}
    t_\emph{Ours} =  \underset{\param{} \in \interval{\param{}}}{\max}~~\ubmu \param{} + \ubnu \quad \textrm{s.t. } \ubalpha \param{} + \ubbeta \geq 0. \label{eq: ourabs_t1}
\end{align}
CROWN-IBP computes the upper bound of $\lossrobust$ as $ \abscibp{\lossrobust}(\x, \cfx{\x}, \param{f}) = \lossmse(\sigma(t_\emph{CIBP}), 0)$, where $t_\emph{CIBP}$ is obtained in the following optimization problem,
\begin{align}
    t_\emph{CIBP} =  \underset{\param{} \in \interval{\param{}}}{\max}~~\ubmu \param{} + \ubnu. \label{eq: cibp_t1}
\end{align}
As \cref{eq: ourabs_t1} has an additional constraint than \cref{eq: cibp_t1}, we have $t_\emph{CIBP} \ge t_\emph{Ours}$.
Therefore, we have $\lossmse(\sigma(t_\emph{CIBP}), 0) \ge \lossmse(\sigma(t_\emph{Ours}), 0)$, which leads to
\[\abscibp{\lossrobust}(\x, \cfx{\x}, \param{f}) \ge \absours{\lossrobust}(\x, \cfx{\x}, \param{f}).\]
\end{proof}

\section{Additional experimental results}

\subsection{Experimental setup}
\paragraph{Details on datasets}
Details on dataset size and feature composition (number of features, as well as continuous/categorical breakdown) are summarized in \cref{tab:datasets}. We create our own distribution shift for CTG and WHO as follows. CTG originally has three classes (normal, suspect, and pathological). We create a distribution shift by first training on only the ``normal'' and ``suspect'' samples. Then, we add in the "pathological" samples (but assign them the same label as the ``suspect'' samples). WHO contains data about countries' life expectancies across several years (2000-2015). We create a temporal data shift by using pre-2012 data as the original dataset, and all data as the shifted dataset. When preprocessing WHO, we also binarize the outcome variable. Instead of predicting the exact life expectancy in years, we predict whether or not it is greater than the pre-2012 median life expectancy.

\begin{table}[t]
    \small
    \centering
    \caption{Dataset composition}
    \begin{tabular}{c|cccc}\toprule
        Dataset & Size (orig.) & Size (shifted) & \# cont. feat. & \# cat. feat. \\\midrule
        HELOC & 9871 & - & 22 & 0 \\
        TC & 30000 & - & 14 & 9 \\
        \student & 32593 & - & 22 & 8 \\
        CTG & 1950 & 2126 & 22 & 0 \\
        WHO & 2196 & 2928 &  18 & 0 \\\bottomrule
    \end{tabular}
    \label{tab:datasets}
\end{table}

\paragraph{Choosing $\delta$}
We dynamically determine a distinct $\delta_i$ value for the $i$-th layer with parameter $\theta_i$ based on its $l_p$ norm, denoted as $\delta_i=\robustratio\|\theta_i\|_{p}$, where $\robustratio \in [0,1]$ is the ratio of $\delta_i$ to the layer norm. 
This design is motivated by two considerations. 
Firstly, parameters across different layers have different $l_p$ norms due to layer size and depth differences. Consequently, a fixed $\delta$ may be overly restrictive for some layers and lenient for others. 
Secondly, using a single fixed $\delta$ could potentially inflate the norm of each layer artificially to enhance $\delta$-Robustness because larger layer norms make the $\delta$-robust multiplicity set relatively smaller under a fixed $\delta$.

\subsection{Class-level data on CTG}
We include additional analysis on class-level performance for CTG. Recall that the original dataset has 3 classes: normal (N), suspect (S), and pathological (P). We train the original model only on N and S data, and then finetune a shifted model with the addition of P data (assigned to the same class as S). We perform this analysis because the dataset is very imbalanced: around 78\% belongs to class N. That is, we want to ensure that the finetuning process actually adapts the model and \cf{} generator to the shifted data (rather than just yielding an overall high performance at the expense of the minority groups S and P). 

Before finetuning, all \cf{s} are valid regardless of class. After finetuning, we find that overall, 98.7\% of \cf{s} generated for CTG remain valid. For samples with class=S (i.e., the minority samples that are represented in the training data), 95.9\% of \cf{s} are valid after finetuning. And for samples with class=P (i.e., only represented in the finetuning data), 96.8\% of \cf{s} remain valid. So, while robustness to finetuning is slightly lower for the minority classes, robustness is very high for all groups. 

\subsection{Ablation Studies}\label{sec:ex-ablation}

\paragraph{Other distance metrics}
In addition to proximity (see \cref{sec:ex-setup} for a definition), we consider \emph{sparsity} and \emph{distance to the data manifold (DDM)} as \cf{} quality metrics. 
Sparsity is defined as the fraction of features that change, i.e., given $\x$ and its $\cfx{\x}$, $\textsc{SPARS}(\x,\cfx{\x}) = \frac{1}{d} \sum_i^d \mathbb{1} [\x_i\neq \cfx{\x}_i]$. 
We approximate DDM by taking the distance to the nearest point in the training data (by using $k$-nearest neighbors with $k=1$). 
For all three metrics, a lower score indicates smaller distance, which indicates that the \cf{} is of higher quality because it will be easier for the user to obtain.  
\Cref{tab:distance} contains data on sparsity and DDM; see \cref{tab:proximity} in the main paper for proximity data. 

\begin{table}[th]
\small
    \centering
    \caption{\small Average sparsity and distance to the data manifold (DDM) for valid \cf{}. (*) indicates that the \cf{} technique does not correctly account for categorical features.} 
    
    \begin{tabular}{ll|lllll}\toprule
   & & CTG  & WHO & HELOC & TC & OULA \\\midrule
\multirow{4}*{Proximity} & \technique{} & 0.780 & 0.758 & 0.817 & 0.758 & 0.424 \\
& CN & 0.765 & 0.718 & 0.791 & 0.751 & 0.403 \\
& ROAR & 0.900 & 0.326 & 0.286 &  0.166* & 0.399*  \\
& SNS &  0.585 & 0.338 & 0.541 & 0.210* & 0.292*
      \\\midrule
\multirow{4}*{DDM} & \technique{} & 0.130 & 0.103 & 0.050 & 0.123 & 0.051 \\
& CN & 0.109 & 0.064 & 0.049 & 0.108 & 0.039 \\
& ROAR &  0.072 & 0.042 & 0.048 &  0.020* & 0.028*  \\
& SNS &  0.052 & 0.040 & 0.054 & 0.015* & 0.024*
      \\\bottomrule
    \end{tabular}
    \label{tab:distance}
\end{table} 

\paragraph{Relation of our approach's effectiveness and the tightness of the bound.}
We compare the three different techniques from \cref{sec:overapprox_loss}. Using an looser bound such as IBP or CROWN-IBP in place of \ourabsname is advantageous for computation speed, however, the overly loose bounds may hurt the effectiveness. 
For instance, we expect to see lower model accuracy 
and higher \cf{} distance metrics with IBP and CROWN-IBP than with \ourabsname. However, \cref{tab:ablation1} shows that three techniques perform largely similarly.
Note that the $\delta$-Robustness rate is computed post-training using \ourabsname as it achieves tighter over-approximation than the other two approaches.
 
\begin{table}[t]
    \small
    \centering
    \caption{\small Fraction of samples that exhibit \deltarobust and empirical random initialization (RI) robustness, along with test accuracy and \cf{} quality metrics for IBP, CROWN-IBP, and \technique{}. Prox. is proxmity, Spars. is sparsity, and Man. is distance to the data manifold.}
    \begin{tabular}{ll|rr|r|rrr}\toprule
    & & $\Delta$-Rob. & RI & Acc. & Prox. & Spars. & Man. \\\midrule
    \multirow{3}*{HELOC} & IBP & 0.746 & 0.920 (0.079) & 0.742 & 0.10 (0.03) & 0.82 (0.08) & 0.05 (0.01) \\
             & C-IBP & 0.721 & 0.918 (0.064) & 0.742 & 0.10 (0.03) & 0.82 (0.08) & 0.05 (0.01) \\
             & \technique{} & 0.788 & 0.932 (0.028) & 0.738 & 0.10 (0.03) & 0.82 (0.08) & 0.05 (0.01) \\\midrule
    \multirow{3}*{TC} & IBP & 0.902 & 0.862 (0.187) & 0.812 & 0.22 (0.06) & 0.74 (0.09) & 0.10 (0.03) \\
             & C-IBP & 0.872 & 0.900 (0.120) & 0.813 & 0.22 (0.06) & 0.75 (0.09) & 0.10 (0.03) \\
             & \technique{} & 0.813 & 0.884 (0.141) & 0.806 & 0.24 (0.05) & 0.76 (0.09) & 0.12 (0.03) \\\midrule
    \multirow{3}*{OULA} & IBP & 0.880 & 0.922 (0.079) & 0.929 & 0.19 (0.04) & 0.45 (0.12) & 0.06 (0.02) \\
             & C-IBP & 0.968 & 0.941 (0.055) & 0.930 & 0.18 (0.04) & 0.43 (0.12) & 0.05 (0.02) \\
             & \technique{} & 0.970 & 0.941 (0.054)  & 0.929 & 0.18 (0.04) & 0.42 (0.12) &  0.05 (0.02)\\\bottomrule
             
    \end{tabular}
    \label{tab:ablation1}
\end{table}


\subsection{Time complexity}
For \technique{}, it takes 8.24 (\textpm 0.14) s. to train one epoch on HELOC, 126.61 (\textpm 1.92) s. to train one epoch on \student, and 27.66 (\textpm 0.36) s. to train one epoch on TC. By contrast, training one epoch with standard training on the same hardware takes 0.55 (\textpm 0.01) s. for HELOC, 5.39 (\textpm 0.06) s. for \student, and 1.80 (\textpm 0.01) s. for TC. We train each dataset for 100 epochs, so with \technique{}, it takes on the order of 14 minutes for HELOC, 3 hours for \student, 45 minutes for TC. With standard training, those times are reduced to around 1 minute, 9 minutes, and 3 minutes, respectively.

\Cref{tab:time} summarizes the time required to generate one \cf{}, given a trained model. (CounterNet is not included, as it takes the same amount of time as \technique{}.) We see that \technique{} can generate \cf{s} near-instantaneously, while ROAR and SNS must solve complex optimization problems for each instance.

\begin{table}[t]
    \small
    \centering
    \caption{\small Average time, in seconds, to generate one counterfactual given a trained model. The standard deviation for \technique{} is negligible ($< 10 ^{-4}$) for all datasets.}
    \begin{tabular}{l|rrr}\toprule
 & HELOC & \student & TC \\\midrule
 \technique{} & $<$0.001 & $<$0.001 & $<$0.001 \\ 
 ROAR & 3.25 \textpm 0.72 & 2.75 \textpm 0.52 & 3.33 \textpm 0.40\\ 
 SNS &  5.25 \textpm 0.21 & 5.92 \textpm 0.21 & 5.89 \textpm 0.20
             
    \end{tabular}
    \label{tab:time}
\end{table} 

\end{document}